\documentclass[10pt,twocolumn,letterpaper]{article}

\usepackage{cvpr}              

\usepackage{graphicx}
\usepackage{amsmath}
\usepackage{amssymb}
\usepackage{booktabs}

\usepackage{epsfig}
\usepackage{graphicx}
\usepackage{amsmath}
\usepackage{amssymb}
\usepackage{multirow}
\usepackage{float}
\usepackage{enumerate}
\usepackage{bm}
\usepackage{wrapfig}
\usepackage{epstopdf}
\usepackage{wrapfig}

\usepackage[pagebackref,breaklinks,colorlinks]{hyperref}

\usepackage[capitalize]{cleveref}
\crefname{section}{Sec.}{Secs.}
\Crefname{section}{Section}{Sections}
\Crefname{table}{Table}{Tables}
\crefname{table}{Tab.}{Tabs.}

\def\cvprPaperID{3492} 
\def\confName{CVPR}
\def\confYear{2023}

\begin{document}

\title{Unsupervised Inference of Signed Distance Functions from Single Sparse Point Clouds without Learning Priors}

\author{Chao Chen$^1$ \quad
    Yu-Shen Liu$^1$
    \thanks{The corresponding author is Yu-Shen Liu. This work was supported by National Key R\&D Program of China (2022YFC3800600), the National Natural Science Foundation of China (62272263, 62072268), and in part by Tsinghua-Kuaishou Institute of Future Media Data.} \quad
    Zhizhong Han$^2$ \\
\fontsize{11pt}{\baselineskip}\selectfont{$^1$School of Software, BNRist, Tsinghua University, Beijing, China} \\
\fontsize{11pt}{\baselineskip}\selectfont{$^2$Department of Computer Science, Wayne State University, Detroit, USA} \\
{\tt\small chenchao19@mails.tsinghua.edu.cn} \quad  {\tt\small liuyushen@tsinghua.edu.cn} \quad {\tt\small h312h@wayne.edu}
}

\maketitle

\begin{abstract}
It is vital to infer signed distance functions (SDFs) from 3D point clouds. The latest methods rely on generalizing the priors learned from large scale supervision. However, the learned priors do not generalize well to various geometric variations that are unseen during training, especially for extremely sparse point clouds. To resolve this issue, we present a neural network to directly infer SDFs from single sparse point clouds without using signed distance supervision, learned priors or even normals. Our insight here is to learn surface parameterization and SDFs inference in an end-to-end manner. To make up the sparsity, we leverage parameterized surfaces as a coarse surface sampler to provide many coarse surface estimations in training iterations, according to which we mine supervision and our thin plate splines (TPS) based network infers SDFs as smooth functions in a statistical way. Our method significantly improves the generalization ability and accuracy in unseen point clouds. Our experimental results show our advantages over the state-of-the-art methods in surface reconstruction for sparse point clouds under synthetic datasets and real scans.The code is available at \url{https://github.com/chenchao15/NeuralTPS}.
\end{abstract}

\section{Introduction}
\label{sec:intro}
Signed distance functions (SDFs) have been a popular 3D representation that shows impressive performance in various tasks~\cite{wang2022neuris,NeuralPoisson,long2022neuraludf,xu2022point,DBLP:conf/icml/GroppYHAL20,wang2022rangeudf,Atzmon_2020_CVPR,Boulch_2022_CVPR,Zhizhong2021icml,jiang2020lig,Peng2021SAP,zhao2020signagnostic,atzmon2020sald,DBLP:journals/corr/abs-2106-10811,yifan2020isopoints,chaompi2022,Mi_2020_CVPR,Genova:2019:LST,jia2020learning,Liu2021MLS,tang2021sign,Peng2020ECCV,ErlerEtAl:Points2Surf:ECCV:2020,li2023NeAF,Baoruicvpr2023,SHS-Net,LP-DIF}. An SDF describes a signed distance field as a mapping from a coordinate to a signed distance, and represents a surface as a level set of the field. We can learn SDFs from signed distance supervision using coordinate-based neural networks. However, obtaining the signed distance supervision requires continuous surfaces such as water-tight manifolds, hence it is still challenging to infer signed distance supervision from raw point clouds due to the discrete character.

Current methods~\cite{DBLP:journals/corr/abs-1901-06802,Park_2019_CVPR,huang2022neuralgalerkin,VisCovolume,aminie2022,Mi_2020_CVPR,Genova:2019:LST,jia2020learning,Liu2021MLS,tang2021sign,Peng2020ECCV,ErlerEtAl:Points2Surf:ECCV:2020,DBLP:conf/cvpr/LiWLSH22} mainly leverage priors to infer SDFs for point clouds. They learn priors from well established signed distance supervision around point clouds during training, and then generalize the learned priors to infer SDFs for unseen point clouds during testing. Although local priors learned at a part level~\cite{Williams_2019_CVPR,Tretschk2020PatchNets,DBLP:conf/eccv/ChabraLISSLN20,jiang2020lig,Boulch_2022_CVPR,DBLP:conf/cvpr/MaLH22} improve the generalization of global priors learned at a shape level~\cite{Mi_2020_CVPR,Genova:2019:LST,jia2020learning,Liu2021MLS,tang2021sign,Peng2020ECCV,ErlerEtAl:Points2Surf:ECCV:2020,sitzmann2019siren}, the geometric variations that local priors can cover are still limited. Hence, some methods~\cite{DBLP:conf/icml/GroppYHAL20,Atzmon_2020_CVPR,zhao2020signagnostic,atzmon2020sald,DBLP:journals/corr/abs-2106-10811,yifan2020isopoints,Zhizhong2021icml,chibane2020neural} try to directly infer SDFs from single point clouds using various strategies~\cite{Zhizhong2021icml,DBLP:conf/icml/GroppYHAL20,Atzmon_2020_CVPR,zhao2020signagnostic,atzmon2020sald,chaompi2022}. However, they require dense point clouds to assure the inference performance, which drastically limits their performance with sparse point clouds in real scans. Therefore, how to infer SDFs from sparse point clouds to achieve better generalization is still a challenge.

To overcome this challenge, we introduce a neural network to infer SDFs from single sparse point clouds. Our novelty lies in the way of inferring SDFs without signed distance supervision, learned priors or even normals, which significantly improves the generalization ability and accuracy in unseen point clouds. We achieve this by learning surface parameterization and SDF inference in an end-to-end manner using a neural network that overfits a single sparse point cloud. To make up the sparsity, the end-to-end learning turns parameterized surfaces as a coarse surface sampler which produces many coarse surface estimations on the fly to statistically infer the SDF. To target extremely sparse point clouds, we parameterize the surface of a point cloud as a single patch on a 2D plane, where 2D samples can be mapped to 3D points that lead to a coarse surface estimation. We further leverage the estimated coarse surface as a reference to infer the SDF based on thin plate splines (TPS) in the feature space, which produces smooth signed distance fields. Our method can statistically infer the SDFs from the permutation of coarse surfaces in different iterations, which reduces the effect of inaccuracy brought by each single coarse surface. Our method outperforms the latest methods under the widely used benchmarks. Our contributions are listed below.

\begin{enumerate}[i)]
\item We introduce a neural network to infer SDFs from single sparse point clouds without using signed distance supervision, learned priors or even normals.
\item We justify the feasibility of learning surface parameterization and inferring SDFs from sparse point clouds in an end-to-end manner. We provide a novel perspective to use surface parameterization to mine supervision.
\item Our method outperforms the state-of-the-art methods in surface reconstruction for sparse point clouds under the widely used benchmarks.
\end{enumerate}

\section{Related Work}
Neural implicit representations have achieved promising performance in various tasks~\cite{mildenhall2020nerf,Oechsle2021ICCV,handrwr2020,seqxy2seqzeccv2020paper,zhizhongiccv2021matching,zhizhongiccv2021completing,pmpnet++,takikawa2021nglod,DBLP:journals/corr/abs-2105-02788,rematasICML21,Han2019ShapeCaptionerGCacmmm,DBLP:journals/corr/abs-2108-03743}. We can learn neural implicit representations from different supervision including 3D supervision~\cite{DBLP:journals/corr/abs-1901-06802,Park_2019_CVPR,aminie2022,MeschederNetworks,chen2018implicit_decoder}, multi-view~\cite{sitzmann2019srns,DIST2019SDFRcvpr,Jiang2019SDFDiffDRcvpr,prior2019SDFRcvpr,shichenNIPS,DBLP:journals/cgf/WuS20,Volumetric2019SDFRcvpr,lin2020sdfsrn,yariv2020multiview,yariv2021volume,geoneusfu,neuslingjie,Yu2022MonoSDF,yiqunhfSDF,Vicini2022sdf,wang2022neuris}, and point clouds~\cite{Williams_2019_CVPR,liu2020meshing,Mi_2020_CVPR,Genova:2019:LST}. We focus on reviewing works related to point clouds below.

\noindent\textbf{Data-Driven based Methods. }With 3D supervision, most methods adopted data-driven strategy to learn priors, and generalized the learned priors to infer implicit representations for unseen point clouds. Some methods learned global priors~\cite{Mi_2020_CVPR,Genova:2019:LST,jia2020learning,Liu2021MLS,tang2021sign,Peng2020ECCV,ErlerEtAl:Points2Surf:ECCV:2020} at a shape level. To improve the generalization of learned priors, some methods learned local priors~\cite{Williams_2019_CVPR,Tretschk2020PatchNets,DBLP:conf/eccv/ChabraLISSLN20,jiang2020lig,Boulch_2022_CVPR,DBLP:conf/cvpr/MaLH22} at a part or patch level. With the learned priors, we can infer implicit representations for unseen point clouds, and then leverage the marching cubes algorithm~\cite{Lorensen87marchingcubes} to reconstruct surfaces.

These methods rely on a large scale dataset to learn priors while they may not generalize well to unseen point clouds that have large geometric variations from the samples in the large scale dataset.

\noindent\textbf{Overfitting based Methods. }For better generalization, some methods focus on learning implicit functions by overfitting neural networks on single point clouds. These methods introduce novel constraints~\cite{DBLP:conf/icml/GroppYHAL20,Atzmon_2020_CVPR,zhao2020signagnostic,atzmon2020sald,yifan2020isopoints,DBLP:journals/corr/abs-2106-10811}, ways of leveraging gradients~\cite{Zhizhong2021icml,chibane2020neural}, differentiable poisson solver~\cite{Peng2021SAP} or specially designed priors~\cite{DBLP:conf/cvpr/MaLH22,DBLP:conf/cvpr/MaLZH22} to learn signed~\cite{Zhizhong2021icml,DBLP:conf/icml/GroppYHAL20,Atzmon_2020_CVPR,zhao2020signagnostic,atzmon2020sald,chaompi2022} or unsigned distance functions~\cite{chibane2020neural,Zhou2022CAP-UDF}. Although these methods have made great progress without learning priors, they require dense point clouds to infer the distance or occupancy fields around point clouds.

\noindent\textbf{Learning from Sparse Point Clouds. }With sparsity, the gap between points on surfaces makes it hard to accurately infer implicit functions. Some methods learned priors~\cite{Boulch_2022_CVPR,DBLP:conf/cvpr/MaLH22}, and conducted test-stage optimization on unseen sparse point clouds~\cite{DBLP:conf/cvpr/MaLH22}. Without priors, NeedleDrop~\cite{Needle3DPoints} was proposed to infer occupancy fields by learning whether a dropped needle goes across the surface or not. However, this self-supervision is not accurate at any point on a surface and heavily relies on the length of needle. VIPSS~\cite{huang2019variational} learns an implicit function from an unoriented point set based on Hermite interpolation, which is sensitive to parameter settings.

Our method falls in this category, but we aim to infer SDFs without learning priors or supervision. We achieve this by learning surface parameterization and SDFs inference in an end-to-end manner for capture a better sense on surfaces.

\noindent\textbf{Neural Splines. }Splines have been widely used in image manipulation~\cite{9880299jian} or generation~\cite{DBLP:conf/nips/DurkanB0P19}. NeuralSpline was proposed to fit point clouds with normals using implicit functions~\cite{DBLP:conf/cvpr/WilliamsTBZ21}. With normals, it simply infers occupancy of points on the normals, hence it focuses on fitting rather than inference. Instead, we target a more challenging scenario where we focus on inferring SDF without normals or learned priors.

\section{Method}
\begin{figure}[tb]
\vspace{-0.2in}
  \centering
   \includegraphics[width=\linewidth]{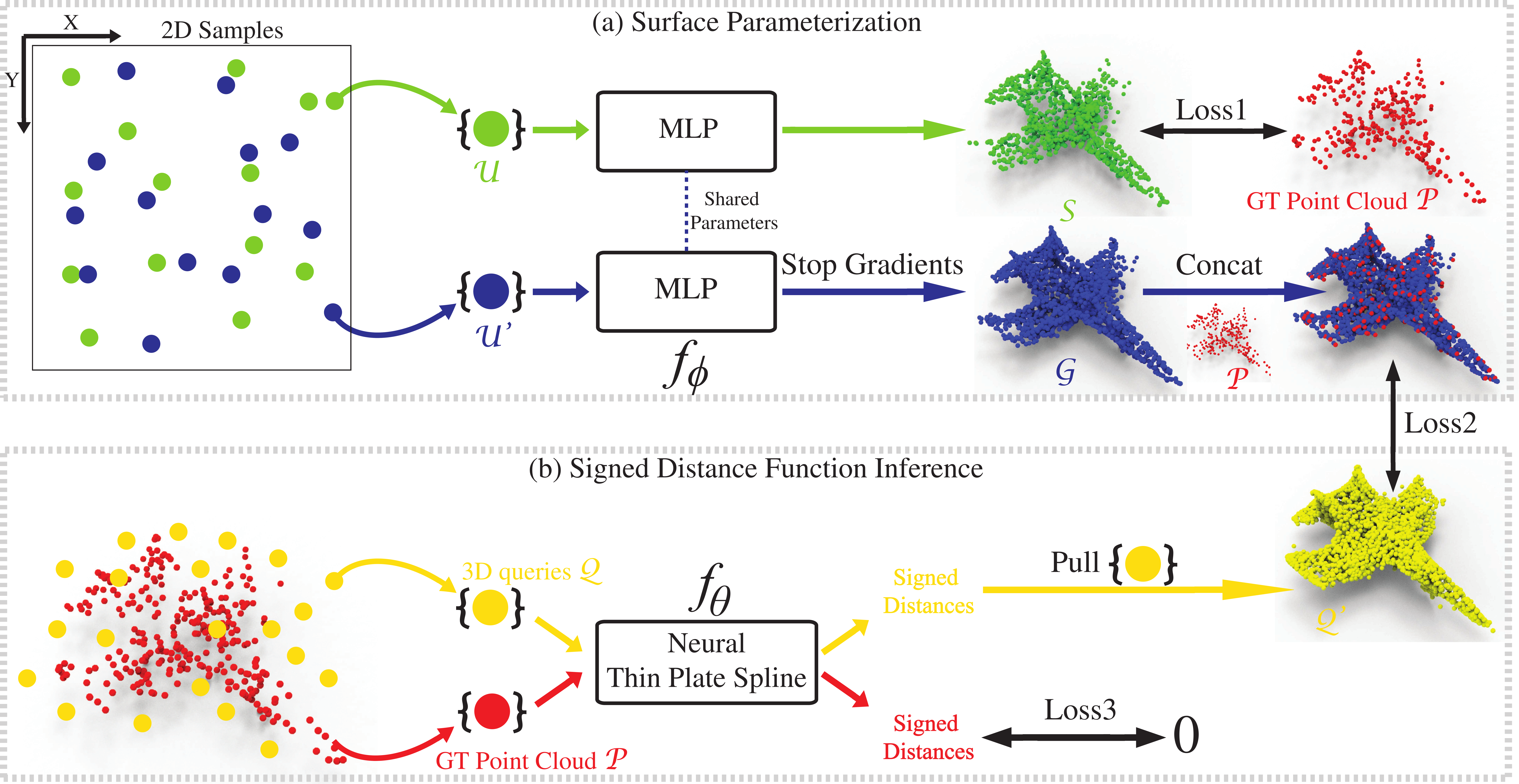}
  %
  %
  \vspace{-0.3in}
\caption{\label{fig:overview}The overview of our method.
}
\vspace{-0.1in}
\end{figure}

\noindent\textbf{Overview. }We aim to infer an SDF $f_\theta$ from a single sparse point cloud $\mathcal{P}=\{p_i|i\in[1,I]\}$, where the SDF is parameterized by a network with parameters $\theta$. At any location $q$ in 3D space, SDF $f_\theta$ predicts signed distance $d=f_\theta(q)$. Our method learns surface parameterization and SDF inference in an end-to-end manner, where we aims to use surface parameterization as a coarse surface estimation sampler to provide supervision for SDF inference. as illustrated in Fig.~\ref{fig:overview}.

During surface parameterization in Fig.~\ref{fig:overview} (a), we randomly sample two sets of 2D points $\mathcal{U}$ and $\mathcal{U}'$ in a unit square in each iteration. We map each 2D sample into a 3D point using a neural network $f_\phi$. This mapping leads to two sets of 3D points $\mathcal{S}=\{s_j|j\in[1,J]\}$ and $\mathcal{G}=\{g_k|k\in[1,K]\}$, each of which forms a chart covering the whole shape. We use point cloud $\mathcal{P}$ to regulate $\{s_j\}$, and use $\{g_k|k\in[1,K]\}$ to mine supervision to infer SDF $f_\theta$ in the following.

We leverage a thin plate splines (TPS) based network (NeuralTPS) to infer SDF $f_\theta$ in Fig.~\ref{fig:overview} (b). Our network learns a feature space, where we leverage TPS interpolation to produce features of arbitrary queries $q\in\mathcal{Q}$ in 3D space based on the features of sparse point cloud $\mathcal{P}$. Our network predicts signed distances $f_\theta(q)$ at $q$ from its interpolated feature. We infer SDF $f_\theta$ by minimizing the difference between the surface $\mathcal{Q}'$ produced with the current inferred signed distance field and the chart $\mathcal{G}$ from the parameterized surface.  Moreover, we also regulate $f_\theta$ by constraining sparse point cloud $\mathcal{P}$ to locate on the zero level of the SDF $f_\theta$.

To remedy the inaccuracy in $\mathcal{G}$, we regard $\mathcal{G}$ in each iteration as a sample of coarse surface estimation, and infer $f_\theta$ by minimizing the loss expectation in a statistical manner. Moreover, we introduce a confidence weight to consider the confidence of each point in $\mathcal{G}$.

\noindent\textbf{Surface Parameterization. }We learn to parameterize a surface represented by a sparse point cloud $\mathcal{P}$ on a 2D plane in Fig.~\ref{fig:overview} (a). We leverage an MLP $f_\phi$ with five layers to learn a mapping from a 2D sample to a 3D point. This mapping produces a 3D chart $\mathcal{S}$ using a set of randomly sampled 2D points $\mathcal{U}$, $\mathcal{S}=f_\phi(\mathcal{U})$. We regulate this mapping by covering $\mathcal{S}$ onto the ground truth points $\mathcal{P}$, which maximizes the overlapping between $\mathcal{S}$ and $\mathcal{P}$ using a Chamfer Distance (CD) loss,

\vspace{-0.15in}
\begin{equation}
\label{eq:cd}
L_{CD}=\frac{1}{J}\sum_{s\in\mathcal{S}}\min_{p\in\mathcal{P}}||s-p||_2^2+\frac{1}{I}\sum_{p\in\mathcal{P}}\min_{s\in\mathcal{S}}||p-s||_2^2.
\vspace{-0.05in}
\end{equation}

\begin{figure}[tb]
  \centering
   \includegraphics[width=\linewidth]{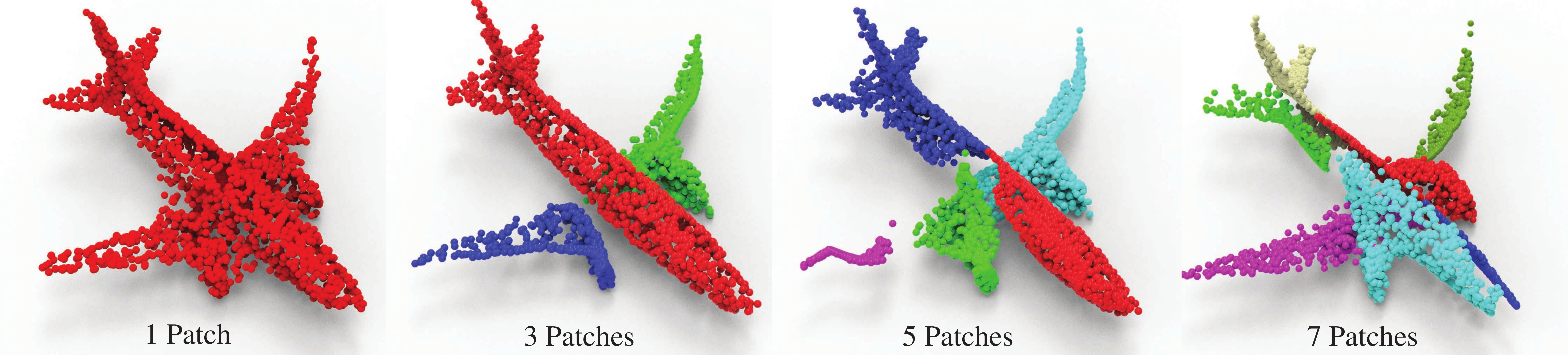}
  %
  %
  \vspace{-0.3in}
\caption{\label{fig:patchnum}The effect of patch numbers.
}
\vspace{-0.2in}
\end{figure}

Our parameterization is similar to AtlasNet~\cite{Groueix_2018_CVPR}. The difference lies in the number of patches to represent a single point cloud. To remedy the sparsity, we only leverage one patch to cover the shape rather than multiple patches in AtlasNet, so that we can better fill the gaps between sparse points using generated points. We visualize the effect of patch numbers in Fig.~\ref{fig:patchnum}, where each subfigure shows a point cloud with the same number of points but different number of patches. The comparison indicates that more patches can not fill the gaps among points, while our single chart reveals a more compact surface.

With surface parameterization, we regard our MLP as a coarse surface sampler which predicts an additional coarse surface estimation $\mathcal{G}$ using another set of 2D samples $\mathcal{U}'$, $\mathcal{G}=f_\phi(\mathcal{U}')$. In each iteration during training, we leverage $\mathcal{G}$ to infer SDF $f_\theta$ in Fig.~\ref{fig:overview} (b). We stop the gradients that can be back-propagated from the loss on $\mathcal{G}$, which avoids the impact of SDF inference on surface parameterization. This is also the reason why we design a two-branch structure for surface parameterization, which differs our method from AtlasNet a lot.

\noindent\textbf{Signed Distance Function Inference. }We introduce NeuralTPS to infer SDF $f_\theta$ from sparse point cloud $\mathcal{P}$. We sample 3D queries $q$ using a Gaussian function centered at each point in $\mathcal{P}$. The inferred SDF predicts signed distances $f_\theta(p_i)$ and $f_\theta(q)$ at each point $p_i\in\mathcal{P}$ and each sampled query $q$. Here, we impose two different constraints to signed distances $f_\theta(p_i)$ on the surface and signed distances $f_\theta(q)$ in 3D space.

For points $p$ on surface of $\mathcal{P}$, we expect them on the zero level set of SDF $f_\theta$, hence we leverage a MSE loss,

\vspace{-0.05in}
\begin{equation}
\label{eq:MSE}
L_{Surf}=\sum_{p\in\mathcal{P}}(f_\theta(p))^2.
\end{equation}

For points $q\in\mathcal{Q}$ in 3D space, we expect the signed distance field could provide the correct signed distances and gradients which can be used to pull $q$ onto the nearest points on the coarse surface $\mathcal{G}$. Here, we use a pulling operation introduced in~\cite{Zhizhong2021icml} to pull $q$ to $q'$, $q'=q-f_{\theta}(q)\nabla f_{\theta}(q)/||\nabla f_{\theta}(q)||_2$. while, different from~\cite{Zhizhong2021icml}, we introduce a novel confidence-weighted loss to optimize the set $\mathcal{Q}'=\{q'\}$ to cover the coarse surface $\mathcal{G}$ with considering the confidence of each point in $\mathcal{G}$,

\vspace{-0.05in}
\begin{equation}
\label{eq:pull}
L_{Pull}(\mathcal{Q}',\mathcal{G})=\sum_{q'\in\mathcal{Q}',g\in\mathcal{G}}w||q'-g||_2^2.
\end{equation}

\noindent where $g$ is the nearest point of $q$ on $\mathcal{G}$. In practice, we find the nearest point from the union of $\mathcal{G}$ and sparse points $\mathcal{P}$. We use $w$ to model the confidence of each point $g\in\mathcal{G}$ to remedy the inaccuracy in $\mathcal{G}$. $g$ has higher confidence if it is nearer to sparse point cloud $\mathcal{P}$ and vice versa. Hence, we formulate $w$ as, $w=exp(-\delta*||g-p||_2^2)$, where $p$ is the nearest point of $g$ on the sparse point cloud $\mathcal{P}$ and the decay parameter $\delta$ is set to 50 in our experiments.

To further reduce the impact of inaccuracy in $\mathcal{G}$, we minimize $L_{Pull}(\mathcal{Q}',\mathcal{G})$ in a statical manner over $\{\mathcal{G}\}$ obtained in different iterations rather than $\mathcal{G}$ in a single iteration. Hence, we aim to find an optimal $\mathcal{Q}'$ that has the smallest average deviation over $\{\mathcal{G}\}$. We reformulate $L_{Pull}(\mathcal{Q}',\mathcal{G})$ into $\mathbb{E}_{\{\mathcal{G}\}}\{L_{Pull}(\mathcal{Q}',\mathcal{G})\}$.

\noindent\textbf{Loss Function. }We optimize surface parameterization and SDF inference in an end-to-end manner by adjusting parameters $\theta$ and $\phi$ using the following objective function,

\vspace{-0.15in}
\begin{equation}
\label{eq:loss}
\min_{\theta,\phi} L_{CD}+\alpha L_{Surf}+\beta \mathbb{E}_{\{\mathcal{G}\}}\{L_{Pull}(\mathcal{Q}',\mathcal{G})\},
\vspace{-0.05in}
\end{equation}

\noindent where $\alpha$ and $\beta$ are balance weights, and we set $\alpha=0.1$ and $\beta=0.1$ in our experiments.

\noindent\textbf{Neural Thin Plate Splines. }We introduce NeuralTPS to infer SDF as a smooth function. Our key idea is to learn an optimal feature space that can be further mapped to signed distances, where we regress signed distances at queries using features of surface points by TPS interpolation.

\begin{figure}[tb]
  \centering
   \includegraphics[width=\linewidth]{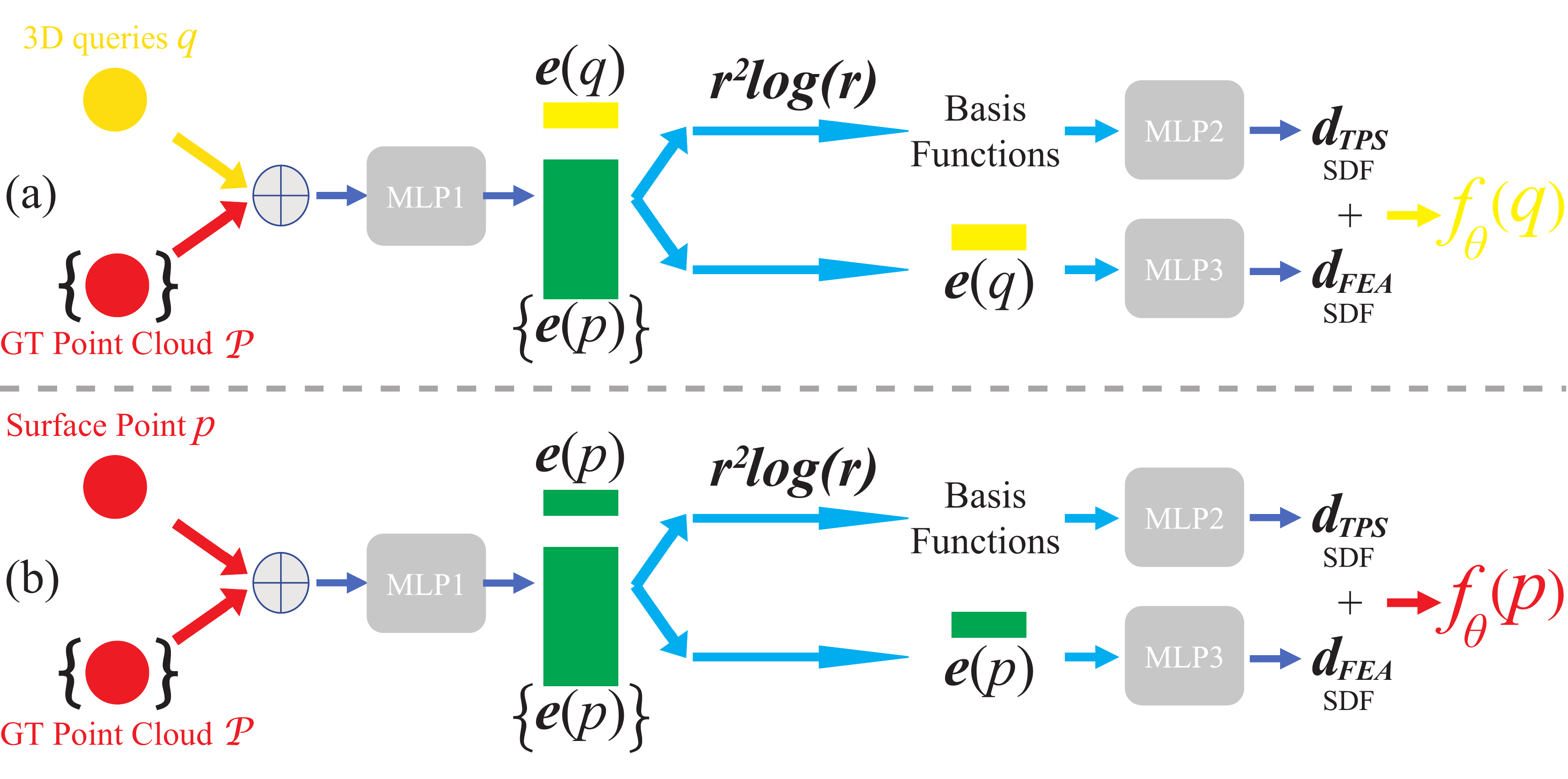}
  %
  %
  \vspace{-0.3in}
\caption{\label{fig:TPS}The illustration of NeuralTPS with one query, such as (a) a surface point or (b) a sampled point.}
\vspace{-0.3in}
\end{figure}

We illustrate NeuralTPS in Fig.~\ref{fig:TPS} (a). We start from concatenating surface point $p_i\in\{\mathcal{P}\}$ with sampled queries $q\in\mathcal{Q}$ in each iteration. This aims to extract point features $\mathbf{e}(p_i)$ and $\mathbf{e}(q)$ using the same parameters for TPS interpolation. So, we leverage an MLP (denoted as MLP1) to learn features of points $p_i$ and query $q$, i.e. $\mathbf{e}(p_i)=MLP(p_i)$ and $\mathbf{e}(q)=MLP(q)$. Then, we regard the features of surface points as control nodes to regress signed distances $d_{TPS}$ at queries $q$ using TPS interpolation below,

\vspace{-0.05in}
\begin{equation}
\label{eq:TPS}
d_{TPS}=\sum_{i=1}^I c_i\psi(||\mathbf{e}(p_i)-\mathbf{e}(q)||_2^2),
\end{equation}

\noindent where $\psi(r)=r^2log(r)$ is known as the thin plate radial basis function, and we will report results with other basis functions in our experiments. $\{c_i\}$ are weights for integrating basis functions, which are learnable parameters in another MLP (denoted as MLP2).

To complement the potential interpolation error in the linear summation, we predict a displacement $d_{FEA}$ for signed distances at queries using point features $\mathbf{e}(q)$ through an MLP (denoted as MLP3). In summary, we formulate signed distances at queries $q$ as,

\vspace{-0.1in}
\begin{equation}
\label{eq:TPS}
f_{\theta}(q)=d_{TPS}+d_{FEA}.
\end{equation}

We use the same way to predict signed distance $f_{\theta}(p)$ of surface points $p\in\{\mathcal{P}\}$, as illustrated in Fig.~\ref{fig:TPS} (b).

\noindent\textbf{Motivation of NeuralTPS. }One of challenge for SDF inference from sparse point clouds is to produce a smooth field. We adopt TPS in the learned feature space, since TPS is a unique solution to scattered data interpolation with maximum smoothness evaluated by second order partial derivatives~\cite{DBLP:journals/tog/TurkO02}. The smoothness is a measurement of the aggregate curvature of $f_{\theta}$ over the region of the surface. Since the smoothness may filter out sharp edges, we conduct TPS interpolation in the learned feature space rather than 3D space.

\noindent\textbf{Details. }In surface parameterization in Fig.~\ref{fig:overview}, our MLP is formed by 5 fully connected layers. In each iteration during training, we sample 2D points to generate $\mathcal{S}$ with $2000$ 3D points to calculate $L_{CD}$ with the ground truth $\mathcal{P}$, and generate $\mathcal{G}$ with $5000$ 3D points to calculate $L_{Pull}$.

In NeuralTPS in Fig.~\ref{fig:TPS}, MLP1 is formed by 10 fully connected layers and other two MLPs are formed by 1 fully connected layer. We establish $\mathcal{Q}$ by sampling queries around each point in $\mathcal{S}$ with a Gaussian distribution, which gets more queries around the surface.

\section{Experiments}
We evaluate our method in surface reconstruction from synthetic point clouds and real scans. The point clouds represent shapes and scenes. For each point cloud, we predict signed distances at grid locations using the inferred SDF $f_{\theta}$, and then run the marching cubes algorithm~\cite{Lorensen87marchingcubes} to extract a surface.

\subsection{Surface Reconstruction For Shapes}
\label{sec:shapes}
\noindent\textbf{Dataset and Metrics. }We evaluate our method in surface reconstruction for rigid shapes and non-rigid shapes in ShapeNet~\cite{shapenet2015} and D-FAUST~\cite{dfaust:CVPR:2017}. We report our evaluations under the test splitting of ShapeNet from NeuralPull~\cite{Zhizhong2021icml} and the test set of D-FAUST. We do not learn priors, and train neural network to overfit to each single point cloud. For each shape, we follow NeedleDrop~\cite{Needle3DPoints} to  randomly sample $300$ points on each shape as the input to each method in evaluations. Using the learned implicit functions, we extract meshes as the reconstructed surfaces. We evaluate the reconstructed surfaces using L1 Chamfer Distance ($CD_{L1}$), L2 Chamfer Distance ($CD_{L2}$), and normal consistency ($NC$), where we sample $100k$ points on the reconstructed surfaces and ground truth surfaces respectively to measure errors.

\noindent\textbf{Evaluations. }We compare our methods with the state-of-the-art methods including NeedleDrop (NDrop)~\cite{Needle3DPoints}, NeuralPull (NPull)~\cite{Zhizhong2021icml}, SAP~\cite{Peng2021SAP}, ShapeGF (ShpGF)~\cite{ShapeGF}, NeuralSplines (NSpline)~\cite{DBLP:conf/cvpr/WilliamsTBZ21}, OnSurf~\cite{DBLP:conf/cvpr/MaLH22}, VIPSS~\cite{huang2019variational}. Here, we do not compare with SAL~\cite{Atzmon_2020_CVPR} or IGR~\cite{DBLP:conf/icml/GroppYHAL20}, since NDrop and NPull showed better performance over them. Except OnSurf, all other methods do not leverage priors during training, and we train all these methods to overfit to the same sparse point clouds separately. We produce the results of OnSurf using its on-surface prior which is trained under a large-scale dataset. To produce the results of ShapeGF, we use PSR~\cite{DBLP:journals/tog/KazhdanH13} to reconstruct meshes from the predicted point clouds, since the code for reconstruction using gradients is not available. For the normals required by NeuralSplines as input, we provide it the normals obtained on the ground truth meshes. We also do not compare the results of NeedleDrop from its original paper, since its official code shows that it samples $300$ points from a mesh in each iteration during training, which is equivalent to observing a much denser point cloud rather than a single sparse point cloud with merely $300$ during training.

\begin{figure*}[tb]
\vspace{-0.2in}
  \centering
   \includegraphics[width=\linewidth]{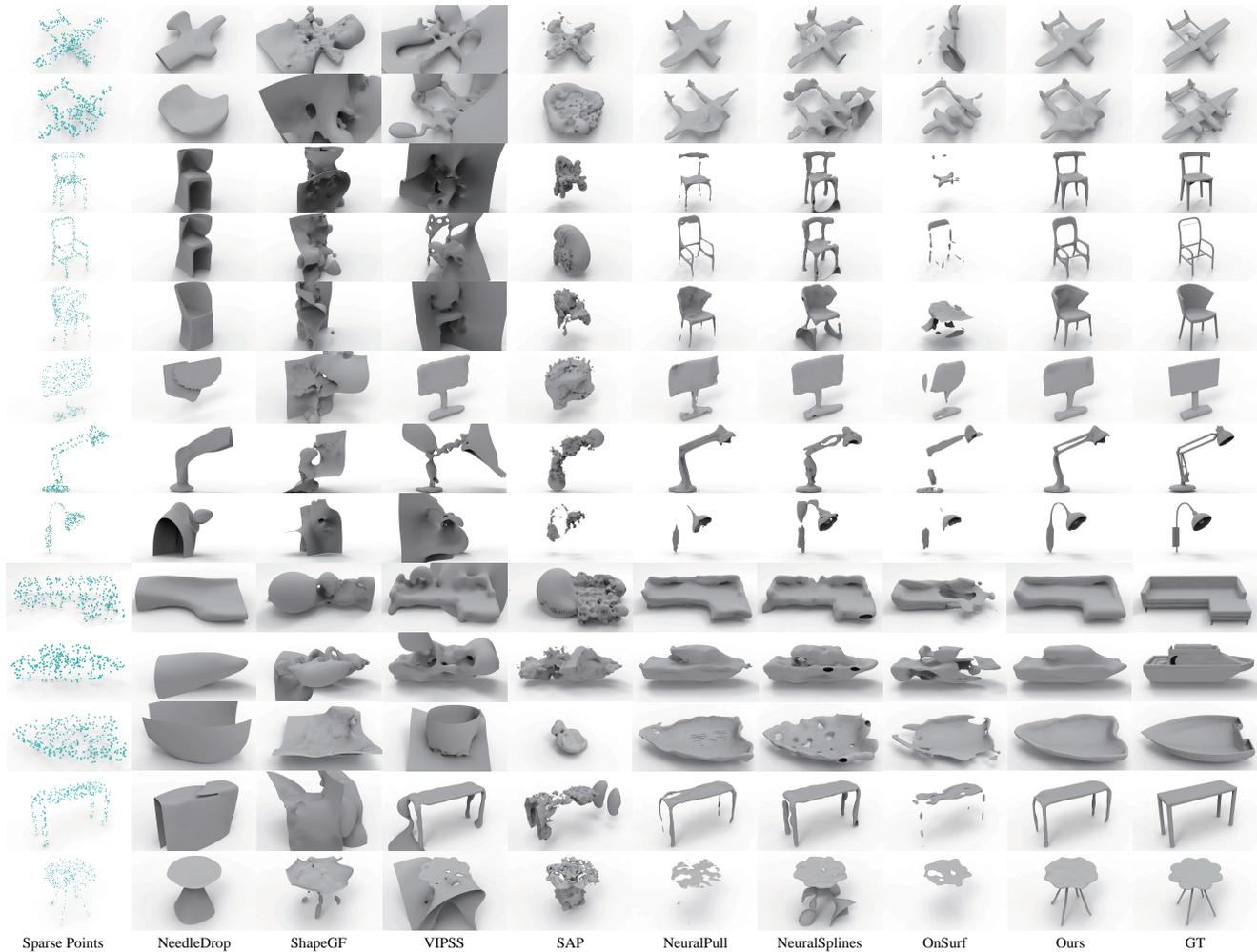}
  %
  %
  \vspace{-0.3in}
\caption{\label{fig:shapenet}Visual comparison with the state-of-the-art under ShapeNet dataset.}
\vspace{-0.1in}
\end{figure*}

We report numerical comparisons in Tab.~\ref{table:NOX1}, Tab.~\ref{table:NOX2}, and Tab.~\ref{table:NOX3}. The comparisons indicate that we achieve the best results which show our superior performance over the latest methods. We also achieve better performance over OnSurf which learns priors for sparse points but does not generalize well to unseen shapes. While methods without learning priors like NeuralPull, NeedleDrop, and SAP can not learn implicit functions from merely $300$ points. Our visual comparisons in Fig.~\ref{fig:shapenet} highlights our advantages in reconstructing more complete and smooth surfaces.

\begin{table}[h]
\vspace{-0.1in}
\centering
\resizebox{\linewidth}{!}{
    \begin{tabular}{c|c|c|c|c|c|c|c|c}
     \hline
      &NDrop&NPull&SAP&ShpGF&NSpline&OnSurf&VIPSS&Ours\\
     \hline
     Plane   &0.499&0.141&0.141&0.110&0.119&0.153&1.193&\textbf{0.095}\\
     Chair   &0.395&0.196&0.363&0.304&0.306&0.316&0.851&\textbf{0.197}\\
     Cabinet &0.229&0.163&0.152&0.604&0.181&0.244&0.584&\textbf{0.138}\\
     Display &0.287&0.145&0.281&0.521&0.193&0.204&0.518&\textbf{0.127}\\
	 Vessel  &0.488&0.116&0.138&0.367&0.134&0.128&0.571&\textbf{0.104}\\
     Table   &0.426&0.400&0.442&0.619&0.318&0.288&1.146&\textbf{0.225}\\
	 Lamp    &0.554&0.162&0.385&0.446&0.231&0.229&0.956&\textbf{0.120}\\
	 Sofa    &0.259&0.139&0.151&0.655&0.168&0.147&0.451&\textbf{0.125}\\
     \hline
     Mean    &0.392&0.183&0.257&0.453&0.206&0.214&0.784&\textbf{0.141}\\
     \hline
   \end{tabular}}
    \vspace{-0.10in}
   \caption{Accuracy of reconstruction with $300$ points under ShapeNet in terms of $CD_{L1}$ $\times$ 10.}  
   \label{table:NOX1}
   \vspace{-0.15in}
\end{table}

\begin{table}[h]
\vspace{-0.15in}
\centering
\resizebox{\linewidth}{!}{
    \begin{tabular}{c|c|c|c|c|c|c|c|c}
     \hline
      &NDrop&NPull&SAP&ShpGF&NSpline&OnSurf&VIPSS&Ours\\
     \hline
     Plane  &0.755&0.036&0.063&0.031&0.127&0.112&5.829&\textbf{0.030}\\
     Chair  &0.532&0.174&0.429&0.275&0.247&0.448&3.291&\textbf{0.149}\\
     Cabinet&0.245&0.086&0.062&0.098&0.064&0.171&2.336&\textbf{0.050}\\
     Display&0.401&0.099&0.311&0.818&0.095&0.153&2.139&\textbf{0.083}\\
	 Vessel &0.844&0.074&0.105&0.439&0.066&0.066&2.614&\textbf{0.051}\\
     Table  &0.701&0.892&0.604&1.117&0.312&0.419&5.009&\textbf{0.272}\\
	 Lamp   &1.071&0.144&0.542&0.591&0.183&0.351&4.617&\textbf{0.051}\\
	 Sofa   &0.463&0.072&0.073&1.253&\textbf{0.053}&0.066&1.890&0.056\\
     \hline
     Mean   &0.627&0.197&0.274&0.578&0.143&0.223&3.47&\textbf{0.093}\\
     \hline
   \end{tabular}}
    \vspace{-0.1in}
   \caption{Accuracy of reconstruction with $300$ points under ShapeNet in terms of $CD_{L2}$ $\times$ 100.}
   \label{table:NOX2}
   \vspace{-0.15in}
\end{table}

\begin{table}[h]
\vspace{-0.1in}
\centering
\resizebox{\linewidth}{!}{
    \begin{tabular}{c|c|c|c|c|c|c|c|c}
     \hline
      &NDrop&NPull&SAP&ShpGF&NSpline&OnSurf&VIPSS&Ours\\
     \hline
     Plane  &0.819&0.897&0.774&0.747&0.895&0.864&0.833&\textbf{0.899}\\
     Chair  &0.777&0.861&0.725&0.547&0.759&0.813&0.821&\textbf{0.863}\\
     Cabinet&0.843&0.888&0.824&0.508&0.840&0.787&0.851&\textbf{0.898}\\
     Display&0.873&0.909&0.744&0.643&0.830&0.855&0.899&\textbf{0.924}\\
	 Vessel &0.838&0.880&0.813&0.667&0.842&0.879&0.867&\textbf{0.908}\\
     Table  &0.795&0.835&0.686&0.601&0.771&0.827&0.783&\textbf{0.877}\\
	 Lamp   &0.828&0.887&0.777&0.673&0.814&0.858&0.848&\textbf{0.902}\\
	 Sofa   &0.808&0.905&0.817&0.508&0.828&0.881&0.882&\textbf{0.919}\\
     \hline
     Mean   &0.823&0.883&0.770&0.612&0.822&0.845&0.848&\textbf{0.899}\\
     \hline
   \end{tabular}}
    \vspace{-0.1in}
   \caption{Accuracy of reconstruction with $300$ points under ShapeNet in terms of $NC$.}
   \label{table:NOX3}
   \vspace{-0.15in}
\end{table}

\begin{figure}[tb]
  \centering
   \includegraphics[width=\linewidth]{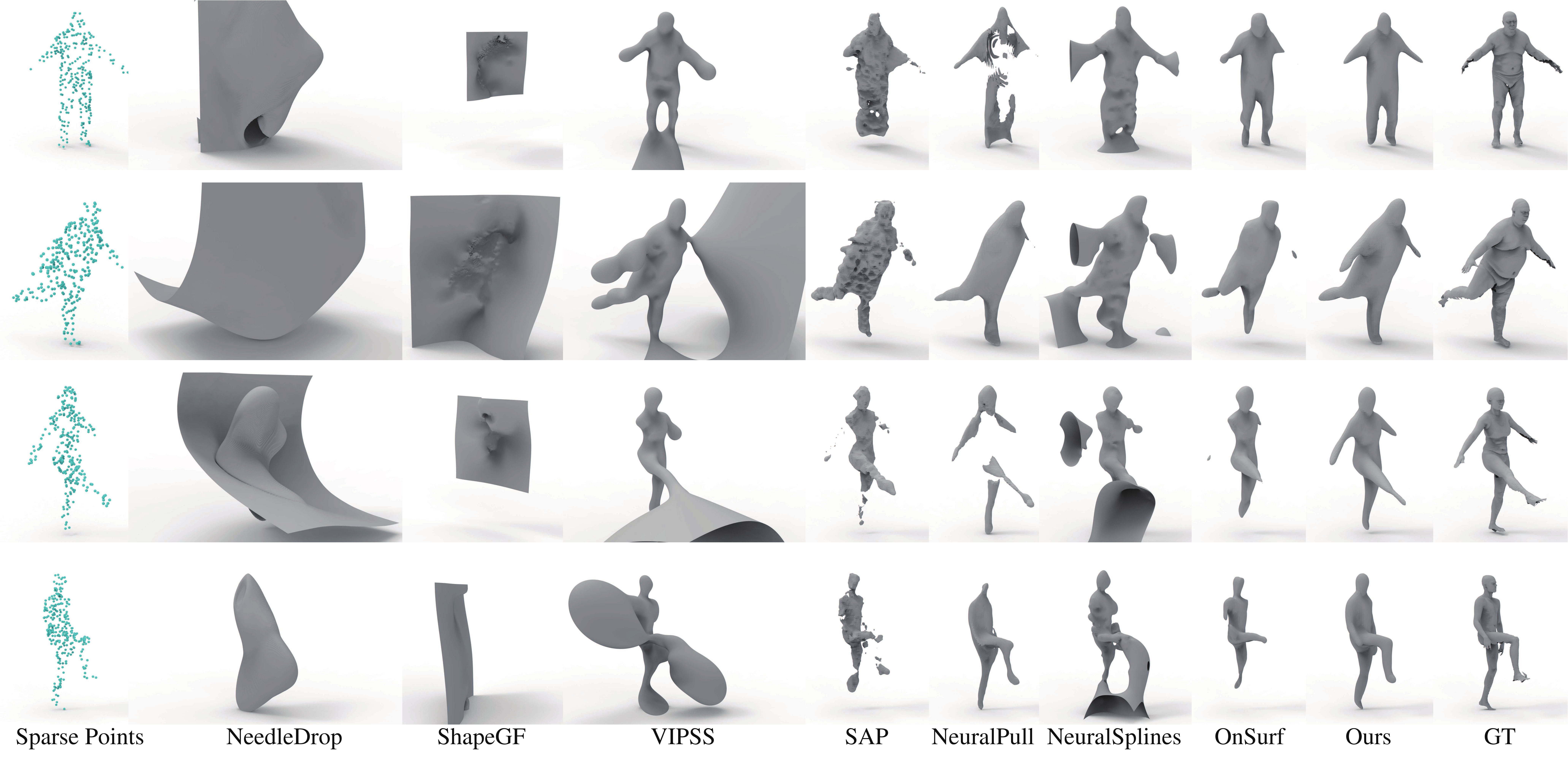}
  %
  %
  \vspace{-0.3in}
\caption{\label{fig:Dfaust}Visual comparison with the state-of-the-art under D-FAUST dataset.}
\vspace{-0.10in}
\end{figure}

We report our evaluations under D-FAUST in Tab.~\ref{table:NOX4}. We follow NeedleDrop to report $CD_{L2}$. We report the $5\%$, $50\%$, and $95\%$ percentiles of the CD between the surface reconstructions and the ground truth. Our method learns better SDFs which achieve better accuracy and smoother surfaces. This is also justified by our visual comparisons in Fig.~\ref{fig:Dfaust}.

\begin{table}[h]
\vspace{-0.10in}
\centering
\resizebox{0.7\linewidth}{!}{
    \begin{tabular}{c|c|c|c|c}
     \hline
     \multirow{2}{*}{Method}&\multicolumn{3}{c|}{$CD_{L2}$$\times$100}&\multirow{2}{*}{NC}\\
     \cline{2-4}
     &5\%&50\%&95\%&\\
     \hline
     NDrop&0.126&1.000&7.404&0.792\\
     NPull&0.018&0.032&0.283&0.877\\
     SAP&0.014&0.024&0.071&0.852\\
     ShpGF&0.452&1.567&8.648&0.750\\
     NSpline&0.037&0.080&0.368&0.808\\
     OnSurf&0.015&0.037&0.123&0.908\\
     VIPSS&0.518&4.327&9.383&0.890\\
    \hline
     Ours&\textbf{0.012}&\textbf{0.160}&\textbf{0.022}&\textbf{0.909}\\
     \hline
   \end{tabular}}
   \vspace{-0.1in}
   \caption{Accuracy of reconstruction with $300$ points under D-FAUST in terms of $CD_{L2}$ and $NC$.}  
   \label{table:NOX4}
   \vspace{-0.3in}
\end{table}

\begin{table*}[thb]
\centering
\resizebox{\linewidth}{!}{
    \begin{tabular}{c|c|c|c|c|c|c|c|c|c|c|c|c|c|c|c|c|c|c}  
     \hline
     \multirow{2}{*}{Method}&\multicolumn{3}{c|}{Burghers}&\multicolumn{3}{c|}{Copyroom}&\multicolumn{3}{c|}{Lounge}&\multicolumn{3}{c|}{Stonewall}&\multicolumn{3}{c|}{Totempole}&\multicolumn{3}{|c}{Mean}\\
     \cline{2-19}
     &$CD_{L1}$&$CD_{L1}$&$NC$&$CD_{L1}$&$CD_{L1}$&$NC$&$CD_{L1}$&$CD_{L1}$&$NC$&$CD_{L1}$&$CD_{L1}$&$NC$&$CD_{L1}$&$CD_{L1}$&$NC$&$CD_{L1}$&$CD_{L1}$&$NC$\\
     \hline
     PSR           &0.178&0.205&0.874&0.225&0.286&0.861&0.280&0.365&0.869&0.300&0.480&0.866&0.588&1.673&0.879&0.314&0.602&0.870\\
     NDrop    &0.200&0.114&0.825&0.168&0.063&0.696&0.156&0.050&0.663&0.150&0.081&0.815&0.203&0.139&0.844&0.175&0.089&0.769\\
     NPull    &0.064&0.008&0.898&0.049&0.005&0.828&0.133&0.038&0.847&0.060&0.005&0.910&0.178&0.024&0.908&0.097&0.016&0.878\\
     SAP           &0.153&0.101&0.807&0.053&0.009&0.771&0.134&0.033&0.813&0.070&0.007&0.867&0.474&0.382&0.725&0.151&0.100&0.797\\
     NSpline &0.135&0.123&0.891&0.056&0.023&0.855&\textbf{0.063}&0.039&0.827&0.124&0.091&0.897&0.378&0.768&0.892&0.151&0.209&0.889\\
     \hline
     Ours&\textbf{0.055}&\textbf{0.005}&\textbf{0.909}&\textbf{0.045}&\textbf{0.003}&\textbf{0.892}&0.129&\textbf{0.022}&\textbf{0.872}&\textbf{0.054}&\textbf{0.004}&\textbf{0.939}&\textbf{0.103}&\textbf{0.017}&\textbf{0.935}&\textbf{0.077}&\textbf{0.010}&\textbf{0.897}\\
     \hline
   \end{tabular}}
   \vspace{-0.1in}
   \caption{Accuracy of reconstruction under 3D Scene in terms of L2CD, L1CD and NC.}  
   \label{table:NOX5}
   \vspace{-0.30in}
\end{table*}

\subsection{Surface Reconstruction For Scenes}
\noindent\textbf{Dataset and Metrics. }We further evaluate our method in surface reconstruction for scenes in 3D Scene~\cite{DBLP:journals/tog/ZhouK13} and KITTI~\cite{Geiger2012CVPR}. For results under 3D scene, we follow previous methods~\cite{Zhizhong2021icml,jiang2020lig} to randomly sample $100$ points per $m^2$. For results under KITTI dataset, we use point clouds in single frames to conduct a comparison. Similarly, we evaluate the reconstructed surfaces using L1 Chamfer Distance ($CD_{L1}$), L2 Chamfer Distance ($CD_{L2}$), and normal consistency ($NC$), where we sample $1000k$ points on the reconstructed surfaces and ground truth surfaces respectively to measure errors.

\begin{figure*}[tb]
\vspace{-0.3in}
  \centering
   \includegraphics[width=\linewidth]{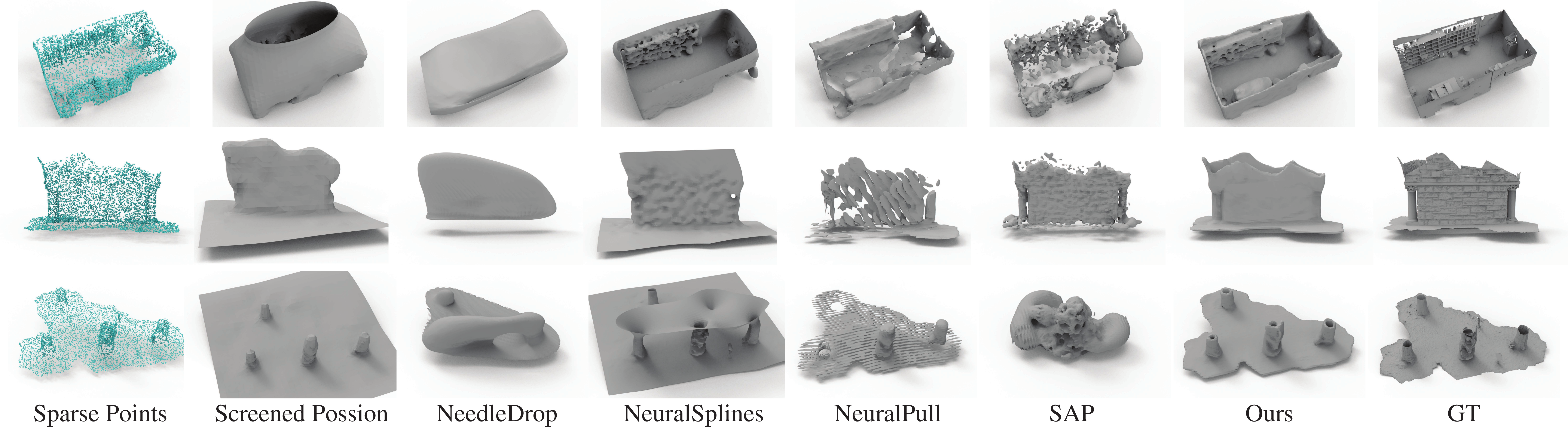}
  %
  %
  \vspace{-0.3in}
\caption{\label{fig:Comp3D}Visual comparison with the state-of-the-art under 3D scene dataset.
}
\vspace{-0.2in}
\end{figure*}

\noindent\textbf{Evaluations. }We compare our methods with the state-of-the-art methods including PSR~\cite{DBLP:journals/tog/KazhdanH13}, NeedleDrop (NDrop)~\cite{Needle3DPoints}, NeuralPull (NPull)~\cite{Zhizhong2021icml}, SAP~\cite{Peng2021SAP}, and NeuralSplines (NSpline)~\cite{DBLP:conf/cvpr/WilliamsTBZ21}. We train each method to overfit each single point cloud. Similarly, we provide NSpline the ground truth normals as input. Our numerical comparisons in Tab.~\ref{table:NOX5} show that our method can reveal more accurate geometry in a 3D scene. Our reconstructed surfaces in Fig.~\ref{fig:Comp3D} are smoother and more complete than others, and do not have artifacts in empty space as NSpline, which justifies our capability of handling sparsity in point clouds.

\begin{figure}[tb]
  \centering
   \includegraphics[width=\linewidth]{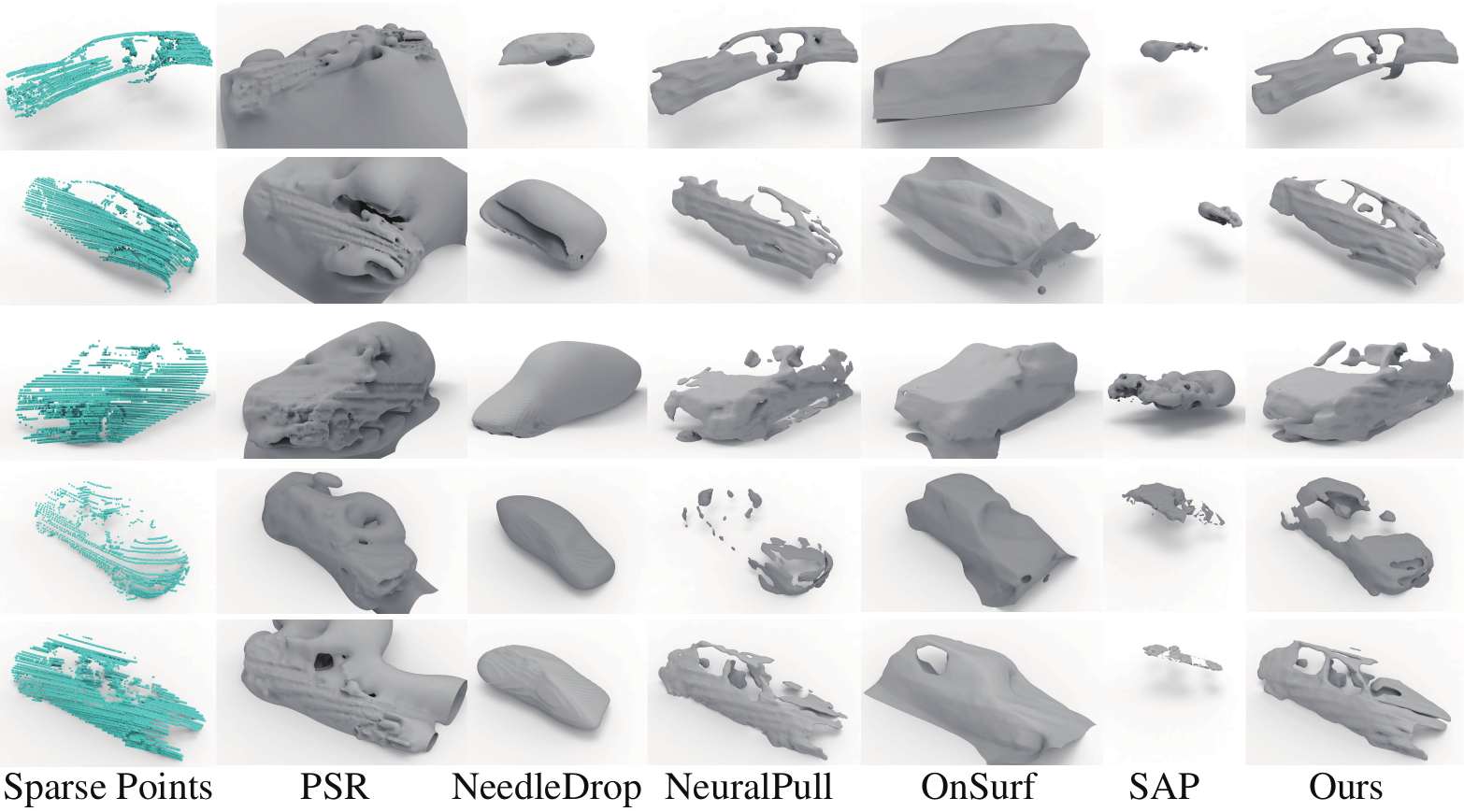}
  %
  %
  \vspace{-0.3in}
\caption{\label{fig:KittiCar}Visual comparisons of cars in KITTI.}
\vspace{-0.3in}
\end{figure}

\begin{figure}[tb]
  \centering
   \includegraphics[width=\linewidth]{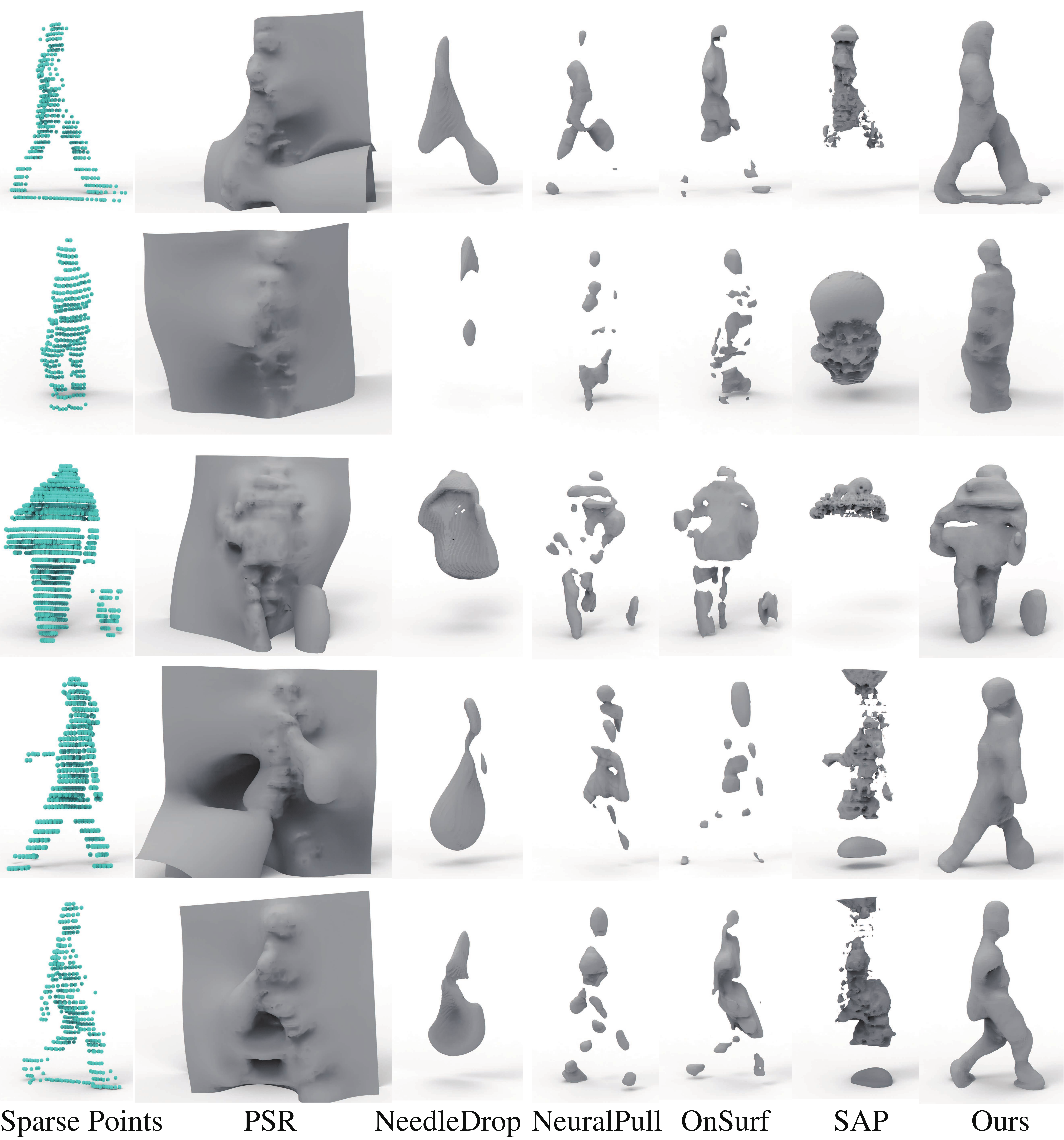}
  %
  %
  \vspace{-0.3in}
\caption{\label{fig:KittiMan}Visual comparisons of pedestrians in KITTI.}
\vspace{-0.1in}
\end{figure}

\begin{figure}[tb]
  \centering
   \includegraphics[width=\linewidth]{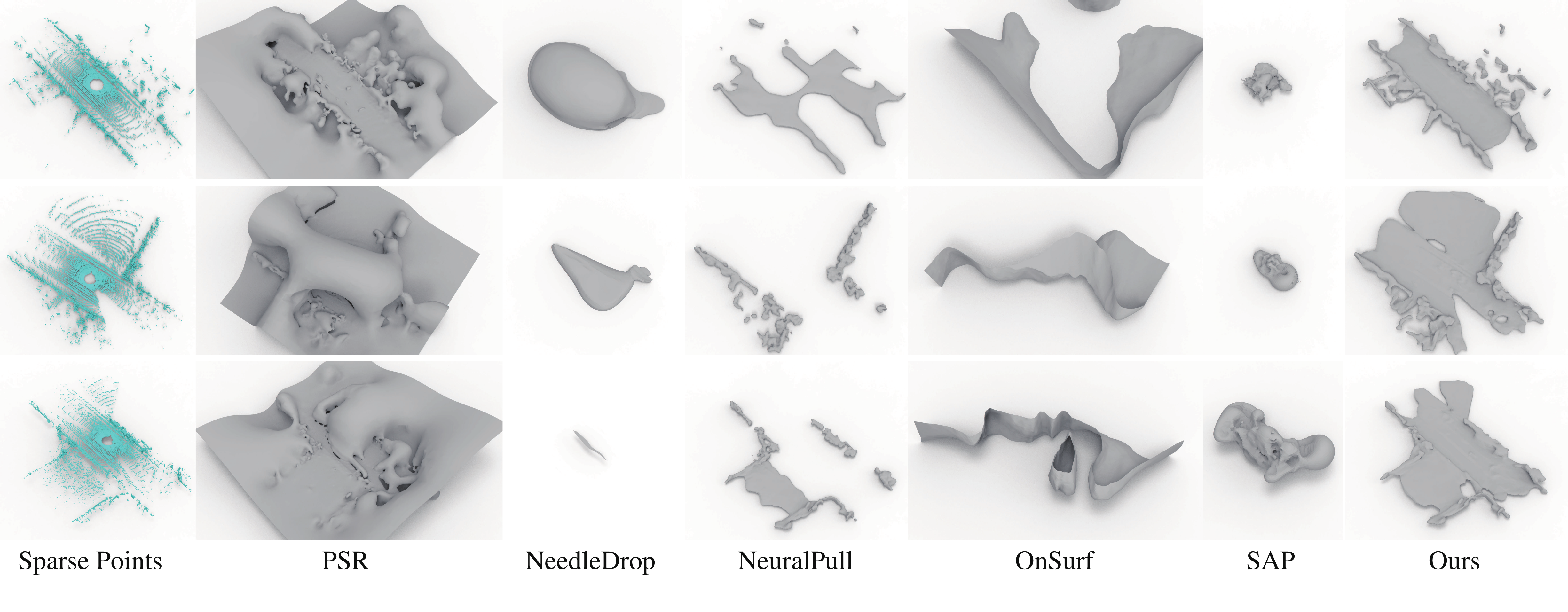}
  %
  %
  \vspace{-0.3in}
\caption{\label{fig:KittiScene}Visual comparisons of roads in KITTI.}
\vspace{-0.2in}
\end{figure}

We further show our reconstructed surfaces from KITTI dataset. Since there are no ground truth meshes, we evaluate our method in visual comparisons with screened possion reconstruction (PSR)~\cite{DBLP:journals/tog/KazhdanH13}, NeedleDrop (NDrop)~\cite{Needle3DPoints}, NeuralPull (NPull)~\cite{Zhizhong2021icml}, SAP~\cite{Peng2021SAP}, and OnSurf~\cite{DBLP:conf/cvpr/MaLH22}. The visual comparisons in reconstructing cars, pedestrians, and roads are shown in Fig.~\ref{fig:KittiCar}, Fig.~\ref{fig:KittiMan} and Fig.~\ref{fig:KittiScene}. Our reconstructed surfaces show more complete surfaces with more geometry details, such as the walking poses of pedestrians, the windows of cars. The smooth roads that we construct highlight our ability of reconstructing thin structures even with sparse points. Our method does not use any learnable priors, and performs much better than the methods without learning priors, such as NeedleDrop and NeuralPull, and also OnSurf which learns a prior for sparse points.

\begin{figure}[tb]
  \centering
   \includegraphics[width=\linewidth]{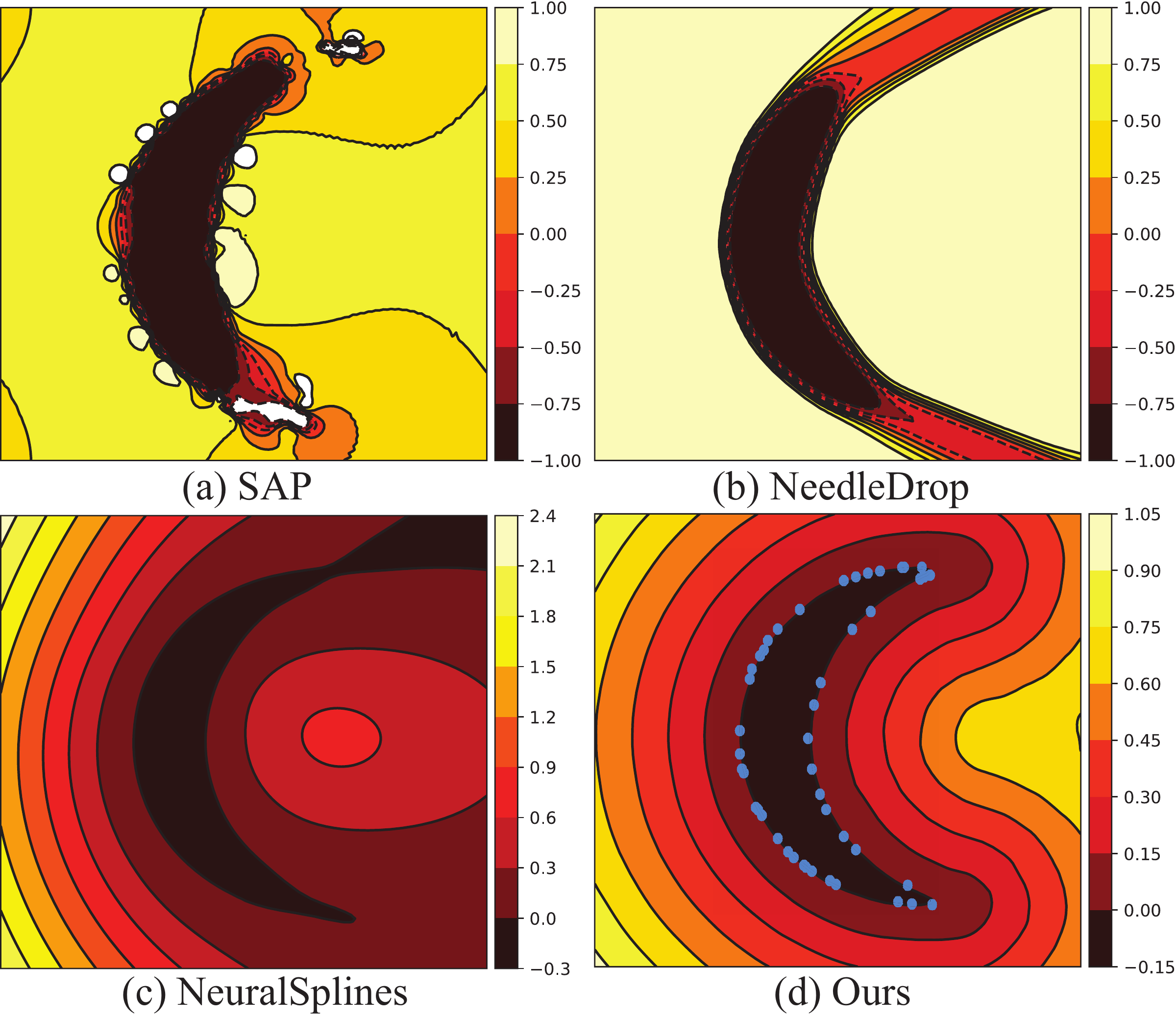}
  %
  %
  \vspace{-0.2in}
\caption{\label{fig:iso}Visual comparison of learned fields with SAP, NeedleDrop, NeuralSplines on a 2D case.
}
\vspace{-0.2in}
\end{figure}

\subsection{Analysis}
We first visualize the signed distance field that our method learns in Fig.~\ref{fig:iso}. To highlight our performance, we conduct visual comparisons with SAP, NeedleDrop, and NeuralSplines on a 2D case. SAP and NeedleDrop estimate occupancy fields, while NeuralSplines and ours learn signed distance fields, and we also provide the ground truth normals to NeuralSplines to produce its results. We use each method to learn an SDF from a sparse 2D point cloud that is nonuniformly sampled on a moon like shape, where we show these points as blue dots in Fig.~\ref{fig:iso}(d). The visual comparisons of level sets learned by each method indicate that our method can employ TPS to reveal smoother level sets with the highest accuracy among the counterparts. Specifically, SAP and NeedleDrop do not deal with sparsity well. Although NeuralSplines also use splines to fit signed distances, it uses the distances along normals as the ground truth, which easily produce artifacts near sharp area.

\subsection{Ablation Studies}
To justify each module of our method, we conduct our ablation studies in airplane class that we used in Sec.~\ref{sec:shapes} under ShapeNet. We report results in surface reconstruction with different variations of our method.

\begin{figure}[tb]
  \centering
   \includegraphics[width=\linewidth]{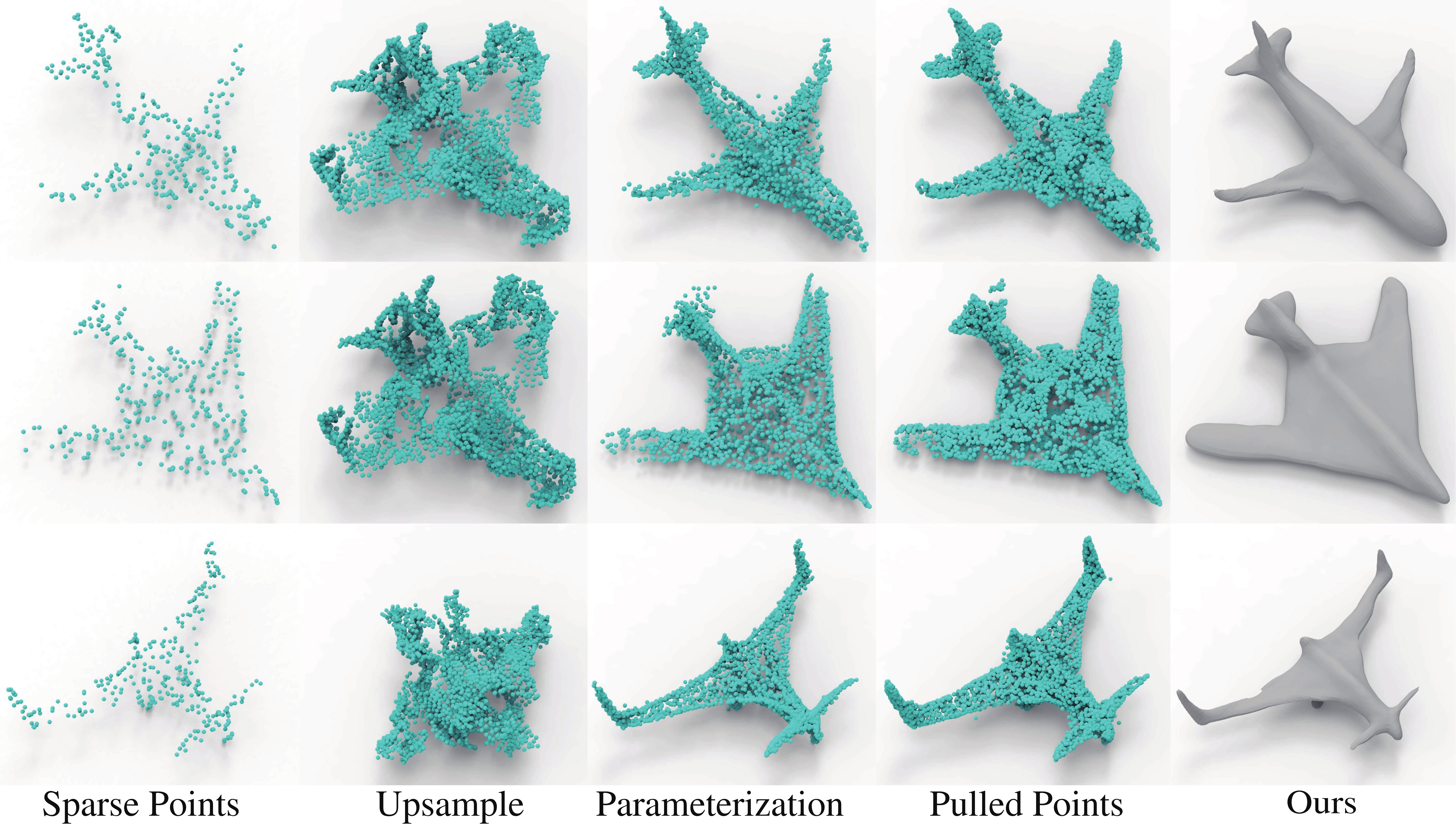}
  %
  %
  \vspace{-0.3in}
\caption{\label{fig:Ablation}Visual comparisons of surface parameterizations.}
\vspace{-0.0in}
\end{figure}

\noindent\textbf{Surface Parameterizations. }We first highlight the benefits of end-to-end training in Tab.~\ref{table:NOXabla1}. Hence, we optimize surface parameterization and SDF inference separately, and merely use the predicted $5000$ points from surface parameterization to infer SDF, as shown by the result of ``Separate''. The degenerated results show that we can infer more accurate SDFs by observing surface estimation in different iteration. Then, we replace surface parameterization into the latest point cloud upsampling method~\cite{feng2022np} to upsample the $300$ point input to $5000$ points. The result of ``Upsample'' indicates that the upsampling method can not generalize the learned prior to upsample $300$ points into a plausible shape with $5000$ points. As shown in Fig.~\ref{fig:Ablation}, the upsampling method is sensitive to the density of input points, resulting in distorted shapes after upsampling. In contrast, both of our surface parameterization and the points pulled to the surface fill the gaps of the sparse point cloud well, which leads to a smooth and complete shape. Next, we show the effect of gradient stop by turning it off. The result of ``GradDiff'' indicates that the error backpropagated from the SDF inference brings too much uncertainty to the surface parameterizations, which turns to degenerate the SDF inference.

\begin{table}[h]
\vspace{-0.0in}
\centering
\resizebox{0.8\linewidth}{!}{
    \begin{tabular}{c|c|c|c|c}
     \hline
     &Separate&Upsample&GradDiff&Ours\\
     \hline
     $CD_{L1}$$\times$10 &0.105&0.582&0.102&\textbf{0.095}\\
     $CD_{L2}$$\times$100&0.118&0.815&0.047&\textbf{0.030}\\
     $NC$             &0.891&0.724&0.897&\textbf{0.899}\\
     \hline
   \end{tabular}}
    \vspace{-0.1in}
   \caption{Effect of surface parameterization.}
   \label{table:NOXabla1}
   \vspace{-0.20in}
\end{table}

We conduct experiments to explore the effect of using more patches for surface parameterizations. As shown in Fig.~\ref{fig:patchnum}, we use more branches like AtlasNet~\cite{Groueix_2018_CVPR} to cover the surface by generating more patches, such as $\{1,3,5\}$. The comparison in Tab.~\ref{table:NOXabla2} shows that more patches degenerate reconstruction accuracy. Since more patches result in larger gaps between patches as we pointed out in Fig.~\ref{fig:patchnum}, which does not resolve the sparsity on the surface.

\begin{table}[h]
\vspace{-0.1in}
\centering
\resizebox{0.6\linewidth}{!}{
    \begin{tabular}{c|c|c|c}
     \hline
     &1&3&5\\
     \hline
     $CD_{L1}$$\times$10 &\textbf{0.095}&0.114&0.115\\
     $CD_{L2}$$\times$100&\textbf{0.030}&0.071&0.077\\
     $NC$             &\textbf{0.899}&0.898&0.896\\
     \hline
   \end{tabular}}
    \vspace{-0.1in}
   \caption{Effect of patch numbers.}
   \label{table:NOXabla2}
   \vspace{-0.15in}
\end{table}

\noindent\textbf{Loss. }We conduct experiments to explore the importance of each term in our loss function in Eq.~\ref{eq:loss}. We remove each of them respectively, and report results in Tab.~\ref{table:NOXloss}. Since we learn SDF, we keep $L_{Pull}$ in all experiments. We first remove $L_{CD}$ and pull queries directly on the sparse points rather than the output of surface parameterization. The degenerated results of ``No $L_{CD}$'' show that the surface parameterization provides an important surface estimation to infer SDFs. Then, we remove $L_{Surf}$, and get slightly worse results of ``No $L_{Surf}$''. These results show the effectiveness of each term, and surface parameterizations supervised by $L_{CD}$ are the most important.

\begin{table}[h]
\vspace{-0.1in}
\centering
\resizebox{0.6\linewidth}{!}{
    \begin{tabular}{c|c|c|c}
     \hline
     &No $L_{CD}$&No $L_{Surf}$&Ours\\
     \hline
     $CD_{L1}$$\times$10 &0.146&0.108&\textbf{0.095}\\
     $CD_{L2}$$\times$100&0.109&0.041&\textbf{0.030}\\
     $NC$             &0.844&0.898&\textbf{0.899}\\
     \hline
   \end{tabular}}
    \vspace{-0.1in}
   \caption{Effect of losses.}
   \label{table:NOXloss}
   \vspace{-0.15in}
\end{table}

\noindent\textbf{Thin Plate Splines. }We report ablation studies related to TPS in Tab.~\ref{table:NOXabla3}. We first replace TPS $\psi(r)=r^2log(r)$ into other splines like $\psi(r)=|r|^3$. The results of ``$|r|^3$'' show that the basis function we use performs better. Then, we highlight the feature space where we do TPS interpolation by removing the MLP1 in Fig.~\ref{fig:TPS}. Instead, we do TPS directly in the spatial space. The result of ``No Feature'' drastically degenerates, which indicates it is more effective to perform TPS interpolation in an optimized feature space than in spatial space. Next, we explore the effect of the displacement $d_{FEA}$ by removing the MLP3 in Fig.~\ref{fig:TPS}. The result of ``No Disp'' shows that predicting SDFs directly from the feature is a good remedy to the TPS prediction.

\begin{table}[h]
\vspace{-0.1in}
\centering
\resizebox{0.8\linewidth}{!}{
    \begin{tabular}{c|c|c|c|c}
     \hline
     &$|r|^3$&No Feature&No Disp&Ours\\
     \hline
     $CD_{L1}$$\times$10 &0.110&0.791&0.159&\textbf{0.095}\\
     $CD_{L2}$$\times$100&0.043&1.699&0.193&\textbf{0.030}\\
     $NC$             &0.895&0.691&0.898&\textbf{0.899}\\
     \hline
   \end{tabular}}
    \vspace{-0.1in}
   \caption{Effect of Thin Plate Splines.}
   \label{table:NOXabla3}
   \vspace{-0.15in}
\end{table}

\begin{figure}[tb]
  \centering
   \includegraphics[width=\linewidth]{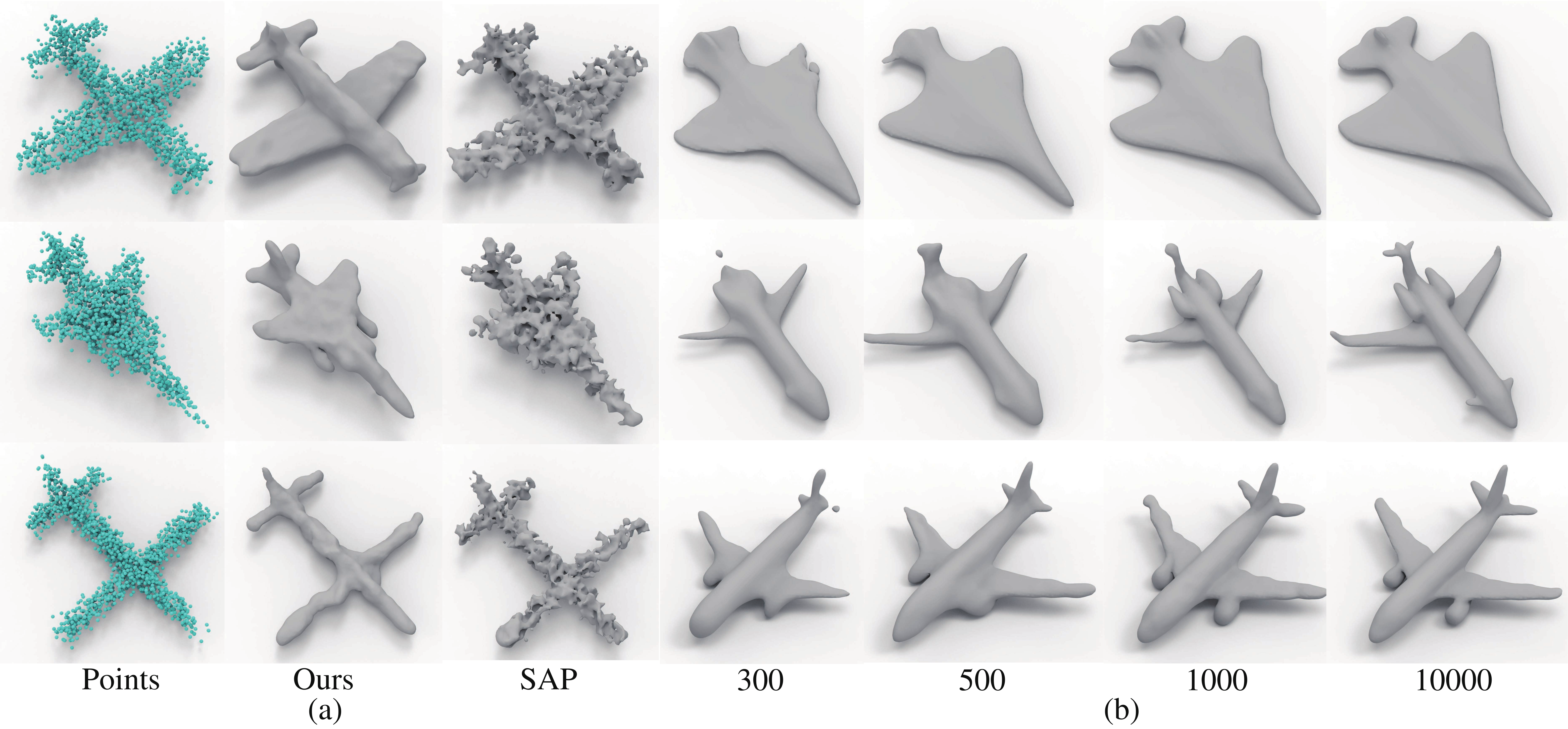}
  %
  %
  \vspace{-0.3in}
\caption{\label{fig:pointnum}(a) Comparisons with noises. (b) Comparisons with different point numbers.}
\vspace{-0.3in}
\end{figure}

\noindent\textbf{Noise Levels. }We report the effect of noises in our method in Tab.~\ref{table:NOXabla4}. We add Gaussian noises with standard deviations including $\{1\%,2\%,3\%\}$. Our results slightly degenerate with $1\%$ and $2\%$ noises and get much worse with $3\%$ noises. Compared to SAP, our method is more robust to noises, as shown in Fig~\ref{fig:pointnum} (a).

\begin{table}[h]
\vspace{-0.10in}
\centering
\resizebox{\linewidth}{!}{
    \begin{tabular}{c|c|c|c|c|c|c|c|c}  
     \hline
     \multirow{2}{*}{}&\multicolumn{4}{c|}{SAP}&\multicolumn{4}{c}{Ours}\\
     \cline{2-9}
     &0\%&1\%&2\%&3\%&0\%&1\%&2\%&3\%\\
     \hline
     $CD_{L1}$$\times$10           &0.141&0.254&0.332&0.561&\textbf{0.095}&0.103&0.177&0.272\\
     $CD_{L2}$$\times$100    &0.063&0.152&0.212&0.348&\textbf{0.030}&0.030&0.041&0.114\\
     $NC$    &0.774&0.645&0.622&0.601&\textbf{0.899}&0.878&0.835&0.803\\
     \hline
   \end{tabular}}
   \vspace{-0.10in}
   \caption{Effect of noise levels.}  
   \label{table:NOXabla4}
   \vspace{-0.15in}
\end{table}

%
%
%
%
%
%

\noindent\textbf{Point Number. }Although our method produces good results on sparse points, we are also able to reconstruct surfaces from dense points, as shown in Fig.~\ref{fig:pointnum} (b).

\section{Conclusion}
We present a method to infer SDFs from single sparse point clouds without using signed distance supervision, learned priors or even normals. We achieve this by learning surface parameterizations and SDF inference in an end-to-end manner. We parameterize the surface as a single chart, which significantly reduces the impact of the sparsity. By evaluating surface parameterization in different iterations, we provide a novel perspective to mine supervision from multiple coarse surface estimations for SDF inference. We also successfully leverage TPS interpolation in feature space to impose smooth constraints on inferring SDF from multiple coarse surface estimations in a statistical way. We justified the effectiveness of key modules and report results that outperform the state-of-the-art methods under the widely used benchmarks.

{\small
\bibliographystyle{ieee_fullname}
\bibliography{papers}
}

\end{document}